\documentclass[conference,compsoc,final]{IEEEtran}
 \IEEEoverridecommandlockouts
\renewcommand\IEEEkeywordsname{Keywords}

\ifCLASSOPTIONcompsoc
  \usepackage[nocompress]{cite}
\else
  \usepackage{cite}
\fi
\ifCLASSINFOpdf
  \usepackage[pdftex]{graphicx}
  \graphicspath{{../pdf/}{../jpeg/}}
  \DeclareGraphicsExtensions{.pdf,.jpeg,.png}
\else
  \usepackage[dvips]{graphicx}
  \graphicspath{{../eps/}}
  \DeclareGraphicsExtensions{.eps}
\fi
\usepackage{amsmath}
\usepackage{algorithmic}
\usepackage{algorithm}
\floatstyle{plaintop}
\restylefloat{algorithm}

\usepackage{array}

\ifCLASSOPTIONcompsoc
  \usepackage[caption=false,font=footnotesize,labelfont=sf,textfont=sf]{subfig}
\else
  \usepackage[caption=false,font=footnotesize]{subfig}
\fi
\usepackage{url}

\usepackage{amssymb}

\begin{document}
%

\title{Deep and Confident Prediction for Time Series  at Uber}



\author{\IEEEauthorblockN{Lingxue Zhu\IEEEauthorrefmark{1}}
\IEEEauthorblockA{
Department of Statistics,\\
Carnegie Mellon University\\
Pittsburgh, Pennsylvania 15213\\
Email: lzhu@cmu.edu}
\thanks{\IEEEauthorrefmark{1} This work was done during an internship at Uber Technologies.}
\and
\IEEEauthorblockN{Nikolay Laptev}
\IEEEauthorblockA{Uber Technologies\\
San Francisco, California 94103\\
Email: nlaptev@uber.com}
}

%


\maketitle

\begin{abstract}
Reliable uncertainty estimation for time series prediction is critical in many fields, including physics, biology, and manufacturing. 
At Uber,
probabilistic time series forecasting is used for robust prediction of number of trips during special events, driver incentive allocation, as well as real-time anomaly detection across millions of metrics. 
Classical time series models are often used in conjunction with a probabilistic formulation for uncertainty estimation. However, such models are hard to tune, scale, and add exogenous variables to.
Motivated by the recent resurgence of Long Short Term Memory networks, we propose a novel end-to-end Bayesian deep model that provides time series prediction along with uncertainty estimation.
We provide detailed experiments of the proposed solution on completed trips data, and successfully apply it to large-scale time series anomaly detection at Uber.
\end{abstract}


%
\IEEEpeerreviewmaketitle


\section{Introduction}

Accurate  time series forecasting and reliable estimation of the prediction uncertainty are critical for anomaly detection, optimal resource allocation, budget planning, and other related tasks. This problem is challenging, especially during high variance segments (e.g., holidays, sporting events), because extreme event prediction depends on numerous external factors that can include weather, city population growth, or marketing changes (e.g., driver incentives) \cite{doi:10.1177/1012690204043462} that all contribute to the uncertainty of the forecast.
These exogenous variables, however, are difficult to incorporate in many classical time series models, such as those found in the standard $R$ \textit{forecast}\cite{forecast} package. In addition, these models usually require manual tuning to set model and uncertainty parameters. 

Relatively recently, time series modeling based on the Long Short Term Memory (LSTM) model \cite{Hochreiter:1997:LSM:1246443.1246450} has gained popularity due to its end-to-end modeling, ease of incorporating exogenous variables, and automatic feature extraction abilities \cite{Assaad:2008:NBA:1297420.1297576}. By providing a large amount of data across numerous dimensions, it has been shown that an LSTM network can model complex nonlinear feature interactions \cite{DBLP:journals/corr/OgunmoluGJG16}, which is critical for modeling complex extreme events. A recent paper \cite{laptev:2017:1273496} has shown that a neural network forecasting model is able to outperform classical time series methods in cases with long, interdependent time series. 

However, the problem of estimating the uncertainty in time-series predictions using neural networks remains an open question. The prediction uncertainty is important for assessing how much to trust the forecast produced by the model, and has profound impact in anomaly detection.
The previous model proposed in \cite{laptev:2017:1273496} had no information regarding the uncertainty. Specifically, this resulted in a large false anomaly rates during holidays where the model prediction has large variance. 

In this paper, we propose a novel end-to-end model architecture for time series prediction, and quantify the prediction uncertainty using Bayesian Neural Network, which is further used for large-scale anomaly detection.

Recently, Bayesian neural networks (BNNs) have garnered increasing attention as a principled framework to provide uncertainty estimation for deep models. 
Under this framework, the prediction uncertainty can be decomposed into three types: {\it model uncertainty}, {\it inherent noise}, and {\it model misspecification}. Model uncertainty, also referred to as epistemic uncertainty, captures our ignorance of the model parameters, and can be reduced as more samples being collected. Inherent noise, on the other hand, captures the uncertainty in the data generation process and is irreducible. These two sources have been previously recognized with successful application in computer visions \cite{kendall2017uncertainties}. 

The third uncertainty from model misspecification, however, has been long-overlooked. This captures the scenario where the testing samples come from a different population than the training set, which is often the case in time series anomaly detection. 
Similar ideas have gained attention in deep learning under the concept of adversarial examples in computer vision \cite{goodfellow2014explaining}, but its implication in prediction uncertainty remains unexplored. Here, we propose a principled solution to incorporate this uncertainty using an encoder-decoder framework. To the best of our knowledge, this is the first time that misspecification uncertainty has been successfully applied to prediction and anomaly detection in a principled way.

In summary, this paper makes the following contributions:
\begin{itemize}
  
  \item Provides a generic and scalable uncertainty estimation implementation for deep prediction models.
  \item Quantifies the prediction uncertainty from three sources: (i) model uncertainty, (ii) inherent noise, and (iii) model misspecification. The third uncertainty has been previously overlooked, and we propose a potential solution with an encoder-decoder.
  \item Motivates a real-world anomaly detection use-case at Uber 
  that uses Bayesian Neural Networks with uncertainty estimation to improve performance at scale.
\end{itemize}

The rest of this paper is organized as follows: Section~\ref{sec:related} gives an overview of previous work on time series prediction for both classical and deep learning models, as well as the various approaches for uncertainty estimation in neural networks. The approach of Monte Carlo dropout (MC dropout) is used in this paper due to its simplicity,  strong generalization ability, and scalability. In Section~\ref{sec:method}, we present our uncertainty estimation algorithm that accounts for the three different sources of uncertainty. Section~\ref{sec:evaluation} provides detailed experiments to evaluate the model performance on 
Uber trip data,
and lays out a successful application to large-scale anomaly detection for millions of metrics at Uber. 
Finally, Section~\ref{sec:conclusion} concludes the paper.


\section{Related Works}
\label{sec:related}

\subsection{Time Series Prediction}

Classical time series models, such as those found in the standard $R$ \textit{forecast}\cite{forecast} package are popular methods to provide an univariate base-level forecast. These models usually require manual tuning to set seasonality and other parameters. Furthermore, while there are time series models that can incorporate exogenous variables \cite{wei1994time}, they suffer from the curse of dimensionality and require frequent retraining. To more effectively deal with exogenous variables, a combination of univariate modeling and a machine learning model to handle residuals was introduced in \cite{2015arXiv150702537O}. The resulting two-stage model, however, is hard to tune, requires manual feature extraction and frequent retraining, which is prohibitive to millions of time series. 

Relatively recently, time series modeling based on LSTM \cite{Hochreiter:1997:LSM:1246443.1246450} technique gained popularity due to its end-to-end modeling, ease of incorporating exogenous variables, and automatic feature extraction abilities \cite{Assaad:2008:NBA:1297420.1297576}. By providing a large amount of data across numerous dimensions, it has been shown that an LSTM approach can model complex extreme events by allowing nonlinear feature interactions \cite{DBLP:journals/corr/OgunmoluGJG16, laptev:2017:1273496}.

While uncertainty estimation for classical forecasting models has been widely studied \cite{622683}, this is not the case for neural networks. Approaches such as a modified loss function or using a collection of heterogenous networks \cite{gal2017concrete} were proposed, however they require changes to the underlying model architecture. A more detailed review is given in the next section. 

In this work, we use a simple and scalable approach for deep model uncertainty estimation that builds on \cite{gal2016dropout}. This framework provides a generic error estimator that runs in production 
at Uber-scale
to mitigate against bad decisions (e.g., false anomaly alerts) 
resulting from poor forecasts due to high prediction variance. 

\subsection{Bayesian Neural Networks}
\label{sec:bnn}
Bayesian Neural Networks (BNNs) introduce uncertainty to deep learning models from a Bayesian perspective. By giving a prior to the network parameters $W$,  the network aims to find the {\it posterior distribution} of $W$, instead of a point estimation.

This procedure is usually referred to as posterior inference in traditional Bayesian models. Unfortunately, due to the complicated non-linearity and non-conjugacy in deep models, exact posterior inference is rarely available; in addition, most traditional algorithms for approximate Bayesian inference cannot scale to the large number of parameters in most neural networks.

Recently, several approximate inference methods are proposed for Bayesian Neural Networks. Most approaches are based on variational inference that optimizes the variational lower bound, including stochastic search \cite{paisley2012variational}, variational Bayes \cite{kingma2013auto}, probabilistic backpropagation \cite{hernandez2015probabilistic}, Bayes by BackProp \cite{blundell2015weight} and its extension \cite{fortunato2017bayesian}. Several algorithms further extend the approximation framework to $\alpha$-divergence optimization, including \cite{hernandez2016black, li2017dropout}. We refer the readers to \cite{gal2016uncertainty} for a more detailed and complete review of these methods.

All of the aforementioned algorithms require different training methods for the neural network. Specifically, the loss function must be adjusted to different optimization problems, and the training algorithm has to be modified in a usually non-trivial sense. In practice, however, an out-of-the-box solution is often preferred, without changing the neural network architecture and can be directly applied to the previously trained model. In addition, most existing inference algorithms introduce extra model parameters, sometimes even double, which is difficult to scale given the large amount of parameters used in practice.

This paper is inspired by the Monte Carlo dropout (MC dropout) framework proposed in \cite{gal2016dropout} and \cite{Gal2015Theoretically}, which requires no change of the existing model architecture and provides uncertainty estimation almost for free. Specifically, stochastic dropouts are applied after each hidden layer, and the model output can be approximately viewed as a random sample generated from the posterior predictive distribution \cite{gal2016uncertainty}. As a result, the model uncertainty can be estimated by the sample variance of the model predictions in a few repetitions. Details of this algorithm will be reviewed in the next section. 

The MC dropout framework is particularly appealing to practitioners because it is generic, easy to implement, and directly applicable to any existing neural networks.  
However, the exploration of its application to real-world problems remains extremely limited. This paper takes an important step forward by successfully adapting this framework to conduct time series prediction and anomaly detection at large scale.


\section{Method}
\label{sec:method}

Given a trained neural network $f^{\hat{W}}(\cdot)$ where $\hat{W}$ represents the fitted parameters, as well as a new sample $x^*$, our goal is to evaluate the uncertainty of the model prediction, $\hat{y}^* = f^{\hat{W}}(x^*)$. Specifically, we would like to quantify the prediction standard error, $\eta$, so that an approximate $\alpha$-level prediction interval can be constructed by 
\begin{equation}
[\hat{y}^* - z_{\alpha/2} \eta, ~ \hat{y}^* + z_{\alpha/2} \eta]
\end{equation}
where $z_{\alpha/2}$ is the upper $\alpha/2$ quantile of a standard Normal. This prediction interval is critical for various tasks. For example, in anomaly detection, anomaly alerts will be fired when the observed value falls outside the constructed 95\% interval. As a result, underestimating $\eta$ will lead to high false positive rates.

In the rest of this section, we will present our uncertainty estimation algorithm in Section~\ref{sec:uncertainty}, which accounts for three different sources of prediction uncertainties. This framework can be generalized to any neural network architectures. Then, in Section~\ref{sec:model-design}, we will present our neural network design for predicting time series at Uber.

\subsection{Prediction Uncertainty}
\label{sec:uncertainty}

We denote a neural network as function $f^W(\cdot)$, where $f$ captures the network architecture, and $W$ is the collection of model parameters. 
In a Bayesian neural network, a prior is introduced for the weight parameters, and the model aims to fit the optimal posterior distribution. For example, a Gaussian prior is commonly assumed:
\[W \sim N(0, I)\]
We further specify the data generating distribution $p(y \,|\, f^W(x))$. In regression, we often assume
\[ y\,|\, W \sim N(f^W(x), \sigma^2) \]
with some noise level $\sigma$. In classification, the softmax likelihood is often used. For time series prediction, we will focus on the regression setting in this paper.

Given a set of $N$ observations $X=\{x_1, ..., x_N\}$ and $Y=\{y_1, ..., y_N\}$, Bayesian inference aims at finding the posterior distribution over model parameters $p(W \,|\, X, Y)$. With a new data point $x^*$, the prediction distribution is obtained by marginalizing out the posterior distribution:
\[ p(y^* \,|\, x^*) = \int_W p(y^* \,|\, f^W(x^*)) p(W \,|\, X, Y)\, dW \]
In particular, the variance of the prediction distribution quantifies the prediction uncertainty, which can be further decomposed using law of total variance:
\begin{equation}
\begin{split}
\textrm{Var}(y^* \,|\, x^*) & = 
\textrm{Var}\left[ \mathbb{E}(y^* \,|\, W, x^*) \right] +
\mathbb{E}\left[\textrm{Var}(y^* \,|\, W, x^*) \right] \\
& = \textrm{Var}(f^W(x^*)) + \sigma^2
\end{split}
\label{eq:var-decompose}
\end{equation}

Immediately, we see that the variance is decomposed into two terms: (i) $\textrm{Var}(f^W(x^*))$, which reflects our ignorance over model parameter $W$, referred to as the {\it model uncertainty}; and (ii) $\sigma^2$ which is the noise level during data generating process,  referred to as the {\it inherent noise}.

An underlying assumption for (\ref{eq:var-decompose}) is that $y^*$ is generated by the same procedure. However, this is not always the case in practice. In anomaly detection, in particular, it is expected that certain time series will have unusual patterns, which can be very different from the trained model. Therefore, we propose that a complete measurement of prediction uncertainty should be a combination from three sources: (i) model uncertainty, (ii) model misspecification, and (iii) inherent noise level. 
The following sections provide details on how we handle these three terms.

\subsubsection{Model uncertainty}
The key to estimating model uncertainty is the posterior distribution $p(W\,|\, X, Y)$, also referred to as Bayesian inference. This is particularly challenging in neural networks because the non-conjugacy due to nonlinearities. There have been various research efforts on approximate inference in deep learning (see Section~\ref{sec:bnn} for a review). Here, we follow the idea in  \cite{gal2016dropout} and \cite{Gal2015Theoretically} to approximate model uncertainty using Monte Carlo dropout (MC dropout). 

The algorithm proceeds as follows: given a new input $x^*$, we compute the neural network output with stochastic dropouts at each layer. That is, randomly dropout each hidden unit with certain probability $p$. This stochastic feedforward is repeated $B$ times, and we obtain $\{\hat{y}^*_{(1)}, ..., \hat{y}^*_{(B)}\}$. Then the model uncertainty can be approximated by the sample variance: 
\begin{equation} 
\widehat{\textrm{Var}}(f^W(x^*)) = \frac{1}{B} \sum_{b=1}^B \left(\hat{y}^*_{(b)} - \overline{\hat{y}}^* \right)^2
\end{equation}
where $\overline{\hat{y}}^* = \frac{1}{B} \sum_{b=1}^B \hat{y}^*_{(b)}$ \cite{gal2016dropout}.
There has been recent work done on choosing the optimal dropout probability $p$ adaptively by treating it as part of the model parameter, but this approach requires modifying the training phase \cite{gal2017concrete}. In practice, we find that the uncertainty estimation is usually robust within a reasonable range of $p$.

\subsubsection{Model misspecification}
Next, we address the problem of capturing potential model misspecification. In particular, we would like to capture the uncertainty when predicting unseen samples with very different patterns from the training data set. We propose to account for this source of uncertainty by introducing an encoder-decoder to the model framework. The idea is to train an encoder that extracts the representative features from a time series, in the sense that a decoder can reconstruct the time series from the encoded space. At test time, the quality of encoding of each sample will provide insight on how close it is to the training set. Another way to think of this approach is that we first fit a latent embedding space for all training time series using an encoder-decoder framework. Then, we measure the distance between test cases and training samples in the embedded space.

The next question is how to incorporate this uncertainty in the variance calculation. Here, we take a principled approach by connecting the encoder, $g(\cdot)$, with a prediction network, $h(\cdot)$, and treat them as one large network $f = h(g(\cdot))$ during inference. \figurename~\ref{fig:model} illustrates such an inference network, and Algorithm~\ref{algo:dropout} presents the MC dropout algorithm. Specifically, given an input time series $x = \{x_1, ..., x_T\}$, the encoder $g(\cdot)$ constructs the learned embedding $e = g(x)$, which is further concatenated with external features, and the final vector is fed to the final prediction network $h$. During this feedforward pass, MC dropout is applied to all layers in both the encoder $g$ and the prediction network $h$. As a result, the random dropout in the encoder perturbs the input intelligently in the embedding space, which accounts for potential model misspecification and gets further propagated through the prediction network. Here, variational dropout for recurrent neural networks \cite{Gal2015Theoretically} is applied to the LSTM layers in the encoder, and regular dropout \cite{gal2016dropout} is applied to the prediction network. 

\begin{algorithm}[H]

 \begin{algorithmic}[1]
 \renewcommand{\algorithmicrequire}{\textbf{Input:}}
 \renewcommand{\algorithmicensure}{\textbf{Output:}}
 \REQUIRE data $x^*$, encoder $g(\cdot)$, prediction network $h(\cdot)$, dropout probability $p$, number of iterations $B$
 \ENSURE prediction $\hat{y}^*_{mc}$, uncertainty $\eta_1$
 \\
 \vspace{3pt}

  \FOR {$b = 1$ to $B$}
  \STATE $e^*_{(b)} \leftarrow$ {\it VariationalDropout}$(g(x^*), p)$
  \STATE $z^*_{(b)} \leftarrow \textrm{Concatenate}(e^*_{(b)}, \textrm{extFeatures})$
  \STATE $\hat{y}^*_{(b)} \leftarrow $  {\it Dropout} $(h(z^*_{(b)}), p)$
  \ENDFOR
  
  \vspace{3pt}
  \textit{// prediction}
  \STATE $\hat{y}^*_{mc} \leftarrow \frac{1}{B} \sum_{b=1}^B \hat{y}^*_{(b)}$
  
  \vspace{3pt}
  \textit{// model uncertainty and misspecification}
  \STATE $\eta_1^2 \leftarrow  \frac{1}{B} \sum_{b=1}^B (\hat{y}^*_{(b)} - \hat{y}^* )^2$

 \RETURN $\hat{y}^*_{mc}, \, \eta_1$
 \end{algorithmic}
 \caption{MCdropout}
 \label{algo:dropout}
 \end{algorithm}
 
\subsubsection{Inherent noise}
Finally, we estimate the inherent noise level $\sigma^2$. In the original MC dropout algorithm \cite{gal2016dropout}, this parameter is implicitly determined by a prior over the smoothness of $W$. As a result, the model could end up with drastically different estimations of the uncertainty level depending on this pre-specified smoothness (see \cite{gal2016uncertainty}, chapter 4). This dependency is undesirable in anomaly detection, because we want the uncertainty estimation to also have robust frequentist coverage, but it is rarely the case that we would know the correct noise level {\it a priori}.

Here, we propose a simple and adaptive approach that estimates the noise level via the residual sum of squares, evaluated on an independent held-out validation set. Specifically, let $f^{\hat{W}}(\cdot)$ be the fitted model on training data, and $X'=\{x'_1, ..., x'_V\}, Y'=\{y'_1, ..., y'_V\}$ be an independent validation set, then we estimate $\sigma^2$ via
\begin{equation} 
\hat{\sigma}^2 = \frac{1}{V} \sum_{v=1}^V \left( y'_v - f^{\hat{W}}(x'_v) \right)^2 \,.
\end{equation}
Note that $(X', Y')$ are independent from $f^{\hat{W}}(\cdot)$, 
and if we further assume that $f^{\hat{W}}(x'_v)$ is an unbiased estimation of the true model, we have 
\begin{equation}
\begin{split}
\mathbb{E}(\hat{\sigma}^2) &=
\sigma^2 +   \frac{1}{V} \sum_{v=1}^V \mathbb{E} \left[ f^{\hat{W}}(x'_v) - f^{W}(x'_v) \right]^2 \\
&= \sigma^2 +   \textrm{Var}_{\rm TRN}(f^{\hat{W}}(x'_v))  
\end{split}
\end{equation}
where $\textrm{Var}_{\rm TRN}$ is w.r.t the training data, which decreases as the training sample size increases, and $\to 0$ as the training sample size $N \to \infty$. Therefore, $\hat{\sigma}^2$ provides an asymptotically unbiased estimation on the inherent noise level. In the finite sample scenario, it always overestimates the noise level and tends to be more conservative.

The final inference algorithm combines inherent noise estimation with MC dropout, and is presented in Algorithm~\ref{algo:inference}.

\begin{algorithm}[H]

 \begin{algorithmic}[1]
 \renewcommand{\algorithmicrequire}{\textbf{Input:}}
 \renewcommand{\algorithmicensure}{\textbf{Output:}}
 \REQUIRE data $x^*$, encoder $g(\cdot)$, prediction network $h(\cdot)$, dropout probability $p$, number of iterations $B$
 \ENSURE prediction $\hat{y}^*$, predictive uncertainty $\eta$
 \\
 \vspace{3pt}
  \textit{// prediction, model uncertainty and misspecification}
   \STATE $\hat{y}^*, \, \eta_1 \leftarrow$ {\it MCdropout} $(x^*, g, h, p, B)$
   
    \vspace{3pt}
 \textit{// Inherent noise}
  \FOR {$x'_v$ {\bf in} validation set $\{x'_1, ..., x'_V\}$}
  \STATE $\hat{y'}_v \leftarrow h(g(x'_v))$
  \ENDFOR
  \STATE $\eta_2^2 \leftarrow \frac{1}{V} \sum_{v=1}^V \left( \hat{y'}_v - y'_v \right)^2$

\vspace{3pt}
 \textit{// total prediction uncertainty}
 \STATE $\eta \leftarrow \sqrt{\eta_1^2 + \eta_2^2}$
 \RETURN $\hat{y}^*, \, \eta$
 \end{algorithmic}
 \caption{Inference}
 \label{algo:inference}
 \end{algorithm}

\subsection{Model Design}
\label{sec:model-design}

\begin{figure}[!t]
\centering
\includegraphics[width=0.5\textwidth]{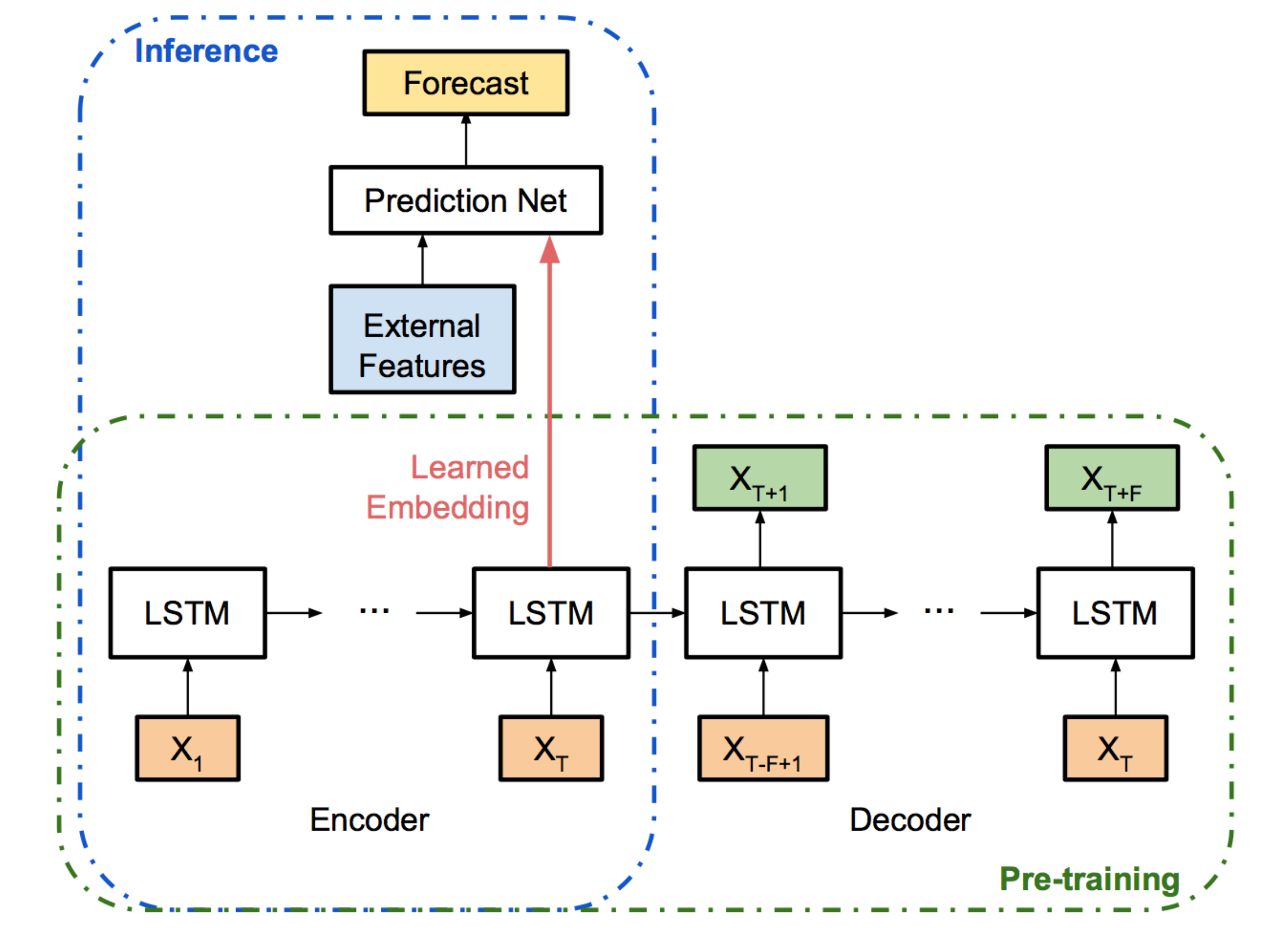}
\caption{Neural network architecture, with a pre-training phase using a LSTM encoder-decoder, followed by a prediction network, with input being the learned embedding concatenated with external features.}
\label{fig:model}
\end{figure}

The complete architecture of the neural network is shown in \figurename~\ref{fig:model}. The network contains two major components: (i) an encoder-decoder framework that captures the inherent pattern in the time series, which is learned during pre-training step, and (ii) a prediction network that takes input from both the learned embedding from encoder-decoder, as well as any potential external features to guide the prediction. We discuss the two components in more details below.

\subsubsection{Encoder-decoder}
Prior to fitting the prediction model, we first conduct a pre-training step to fit an encoder that can extract useful and representative embeddings from a time series. The goals are to ensure that (i) the learned embedding provides useful features for prediction and (ii) unusual input can be captured in the embedded space, which will get further propagated to the prediction network in the next step. 
Here, we use an encoder-decoder framework with two-layer LSTM cells. 

Specifically, given a univariate time series $\{x_t\}_t$, the encoder reads in the first $T$ timestamps $\{x_1, ..., x_T\}$, and constructs a fixed-dimensional embedding state. After then, from this embedding state, the decoder constructs the following $F$ timestamps $\{x_{T+1}, ..., x_{T+F}\}$ with guidance from $\{x_{T-F+1}, ..., x_T\}$ (\figurename~\ref{fig:model}, bottom panel).
The intuition is that in order to construct the next few timestamps, the embedding state must extract representative and meaningful features from the input time series.
This design is inspired from the success of video representation learning using a similar architecture \cite{srivastava2015unsupervised}.

\subsubsection{Prediction network}
After the encoder-decoder is pre-trained, it is treated as an intelligent feature-extraction blackbox. Specifically, the last LSTM cell states of the encoder are extracted as learned embedding. Then, a prediction network is trained to forecast  the next one or more timestamps using the learned embedding as features. In the scenario where external features are available, these can be concatenated to the embedding vector and passed together to the final prediction network.

Here, we use a multi-layer perceptron as the prediction network. We will show in Section~\ref{sec:trip} that the learned embedding from the encoder successfully captures interesting patterns from the input time series. In addition, including external features significantly improves the prediction accuracy during holidays and special events (see Section~\ref{sec:evaluation})

\subsubsection{Inference}

After the full model is trained, the inference stage involves only the encoder and the prediction network (\figurename~\ref{fig:model}, left panel). The complete inference algorithm is presented in Algorithm~\ref{algo:inference}, where the prediction uncertainty, $\eta$, contains two terms: (i) the model and misspecification uncertainty, estimated by applying MC dropout to both the encoder and the prediction network, as presented in Algorithm~\ref{algo:dropout}; and (ii) the inherent noise level, estimated by the residuals on a held-out validation set. Finally, an approximate $\alpha$-level prediction interval is constructed by $[\hat{y}^* - z_{\alpha/2} \eta, ~ \hat{y}^* + z_{\alpha/2} \eta]$, where $z_{\alpha/2}$ is the upper $\alpha/2$ quantile of a standard Normal.

Two hyper-parameters need to be specified in Algorithm~\ref{algo:inference}: the dropout probability, $p$, and the number of iterations, $B$. As for the dropout probability, we find in our experiments that the uncertainty estimation is relatively stable across a range of $p$, and we choose the one that achieves the best performance on the validation set. As for the number of iterations, the standard error of the estimated prediction uncertainty is proportional to $1/\sqrt{B}$. We measure the standard error across different repetitions, and find that a few hundreds of iterations are usually suffice to achieve a stable estimation.


\section{Evaluation}
\label{sec:evaluation}

This section contains two sets of results. We first evaluate the model performance on a moderately sized data set of daily trips 
processed by the Uber platform.
We will evaluate the prediction accuracy and the quality of uncertain estimation during both holidays and non-holidays. We will also present how the encoder recognizes the day of the week pattern in the embedding space. Next, we will illustrate the application of this model to real-time large-scale anomaly detection for millions of metrics at Uber. 

\subsection{Results on 
Uber 
Trip Data}
\label{sec:trip}

\subsubsection{Experimental settings}
In this section, we illustrate the model performance using the daily completed trips over four years across eight representative large cities in U.S. and Canada, including Atlanta, Boston, Chicago, Los Angeles, New York City,  San Francisco, Toronto, and Washington D.C. We use three years of data as the training set, the following four months as the validation set, and the final eight months as the testing set. The encoder-decoder is constructed with two-layer LSTM cells, with 128 and 32 hidden states, respectively. The prediction network has three fully connected layers with {\it tanh} activation, with 128, 64, and 16 hidden units, respectively.

Samples are constructed using a sliding window with step size one, where each sliding window contains the previous 28 days as input, and aims to forecast the upcoming day. The raw data are log-transformed to alleviate exponential effects. Next, within each sliding window, the first day is subtracted from all values, so that trends are removed and the neural network is trained for the incremental value. At test time, it is straightforward to revert these transformations to obtain predictions at the original scale.

\subsubsection{Prediction performance}

We compare the prediction accuracy among four different models:
\begin{enumerate}
\item {\bf Last-Day}: A naive model that uses the last day's completed trips as the prediction for the next day.
\item {\bf QRF}: Based on the naive last-day prediction, a quantile random forest (QRF) is further trained to estimate the holiday lifts, i.e., the ratio to adjust the forecast during holidays. The final prediction is calculated from the last-day forecast multiplied by the estimated ratio.
\item {\bf LSTM}: A vanilla LSTM model with similar size as our model. Specifically, a two-layer sacked LSTM is constructed, with 128 and 32 hidden states, respectively, followed by a fully connected layer for the final output. This neural network also takes 28 days as input, and predicts the next day.
\item {\bf Our Model}: Our model that combines an encoder-decoder and a prediction network, as described in \figurename~\ref{fig:model}.
\end{enumerate}

Table~\ref{tab:prediction} reports the Symmetric Mean Absolute Percentage Error (SMAPE) of the four models, evaluated on the testing set. We see that using a QRF to adjust for holiday lifts is only slightly better than the naive prediction. On the other hand, a vanilla LSTM neural network provides an average of 26\% improvement across the eight cities. As we further incorporate the encoder-decoder framework and introduce external features for holidays to the prediction network (\figurename~\ref{fig:model}), our proposed model achieves another 36\% improvement in prediction accuracy. Note that when using LSTM and our model, only one generic model is trained, where the neural network is not tuned for any city-specific patterns; nevertheless, we still observe significant improvement on SMAPE across all cities when compared to traditional approaches. 

\begin{table}[!t]
\renewcommand{\arraystretch}{1.3}
\caption{SMAPE of Four Different Prediction Models, Evaluated on the Test Data.}
\label{tab:prediction}
\centering
\begin{tabular}{| c | c | c | c | c |}
\hline
{\bf City} & {\bf Last-Day} & {\bf QRF} & {\bf LSTM} & {\bf Our Model}\\
\hline
Atlanta & 15.9 & 13.2  & 11.0 & 7.3 \\
\hline
Boston & 13.6 & 15.4 & 10.0 & 8.2 \\
\hline
Chicago & 16.0 & 12.7 & 9.5 & 6.1 \\
\hline
Los Angeles & 12.3 & 10.9 & 8.5 & 4.7 \\
\hline
New York City & 11.5 & 10.9 & 8.7 & 6.1 \\
\hline
San Francisco & 10.7 & 11.8 & 7.3 & 4.5 \\
\hline
Toronto & 15.2 & 11.7 & 10.0 & 5.3 \\
\hline
Washington D.C. & 13.0 & 13.3 & 8.2 & 5.2 \\
\hline
{\bf Average} & {\bf 13.5} & {\bf 12.5} & {\bf 9.2} &  {\bf 5.9} \\
\hline
\end{tabular}
\end{table}

\begin{figure}[!t]
\centering
\includegraphics[width=0.5\textwidth]{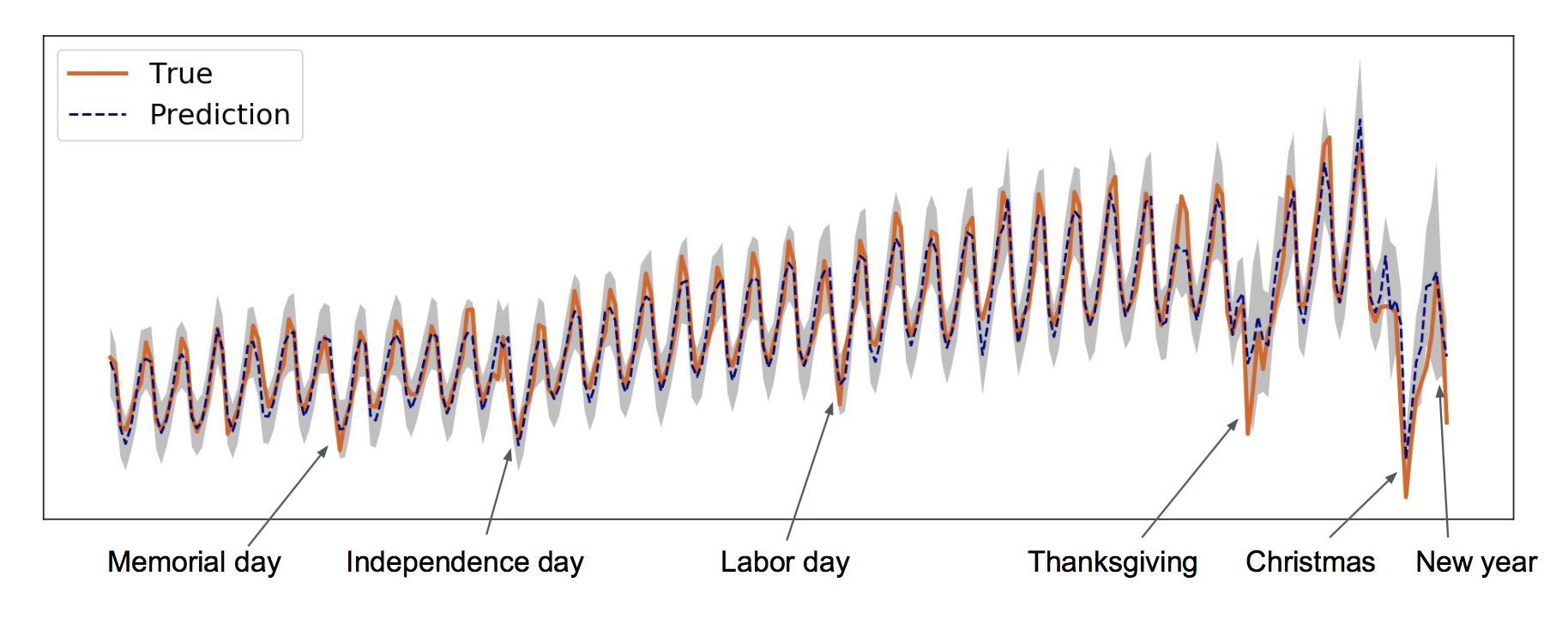}
\caption{Daily completed trips in San Francisco during eight months of the testing set. True values are shown with the orange solid line, and predictions are shown with the blue dashed line, where the 95\% prediction band is shown as the grey area. Exact values are anonymized. }
\label{fig:sf-prediction}
\end{figure}

Finally, \figurename~\ref{fig:sf-prediction} visualizes the true values and our predictions during the testing period in San Francisco as an example. We observe that accurate predictions are achieved not only in regular days, but also during holiday seasons.

\subsubsection{Uncertainty estimation}
Next, we evaluate the quality of our uncertainty estimation by calibrating the empirical coverage of the prediction intervals. Here, the dropout probability is set to be 5\% at each layer, and Table~\ref{tab:coverage} reports the empirical coverage of the 95\% predictive intervals under three different scenarios: 

	\begin{enumerate}
	\item {\bf PredNet}: Use only model uncertainty estimated from MC dropout in the prediction network, with no dropout layers in the encoder.
	\item {\bf Enc+Pred}: Use MC dropout in both the encoder and the prediction network, but without the inherent noise level. This is the term $\eta_1$ in Algorithm~\ref{algo:inference}. 
    \item {\bf Enc+Pred+Noise}: Use the full prediction uncertainty $\eta$ as presented in Algorithm~\ref{algo:inference},  including $\eta_1$ as in 2), as well as the inherent noise level $\eta_2$.
	\end{enumerate}
    
\begin{table}[!t]
\renewcommand{\arraystretch}{1.3}
\caption{Empirical Coverage of 95\% Predictive Intervals, Evaluated on the Test Data.
}
\label{tab:coverage}
\centering
\begin{tabular}{| c | c | c | c |}
\hline
{\bf City} & {\bf PredNet}  & {\bf Enc+Pred} & {\bf Enc+Pred+Noise}\\
\hline
Atlanta & 78.33\% & 91.25\% & 94.30\% \\
\hline
Boston & 85.93\% & 95.82\% & 99.24\% \\
\hline
Chicago & 71.86\% & 80.23\% & 90.49\% \\
\hline
Los Angeles & 76.43\% & 92.40\%  & 94.30\% \\
\hline
New York City & 76.43\% & 85.55\%  & 95.44\% \\
\hline
San Francisco & 78.33\% & 95.06\% & 96.20\% \\
\hline
Toronto & 80.23\% & 90.87\% & 94.68\% \\
\hline
Washington D.C. & 78.33\% & 93.54\% & 96.96\% \\
\hline
{\bf Average} & {\bf 78.23\%} & {\bf 90.59\%} &  {\bf 95.20\%} \\
\hline
\end{tabular}
\end{table}
By comparing {\tt PredNet} with {\tt Enc+Pred}, it is clear that introducing MC dropout to the encoder network is critical, which significantly improves the empirical coverage from 78\% to 90\% by capturing potential model misspecification. In addition, by further accounting for the inherent noise level, the empirical coverage of the final uncertainty estimation, {\tt Enc+Pred+Noise}, nicely centers around 95\% as desired.  

\begin{figure}[!t]
\centering
\includegraphics[width=0.48\textwidth]{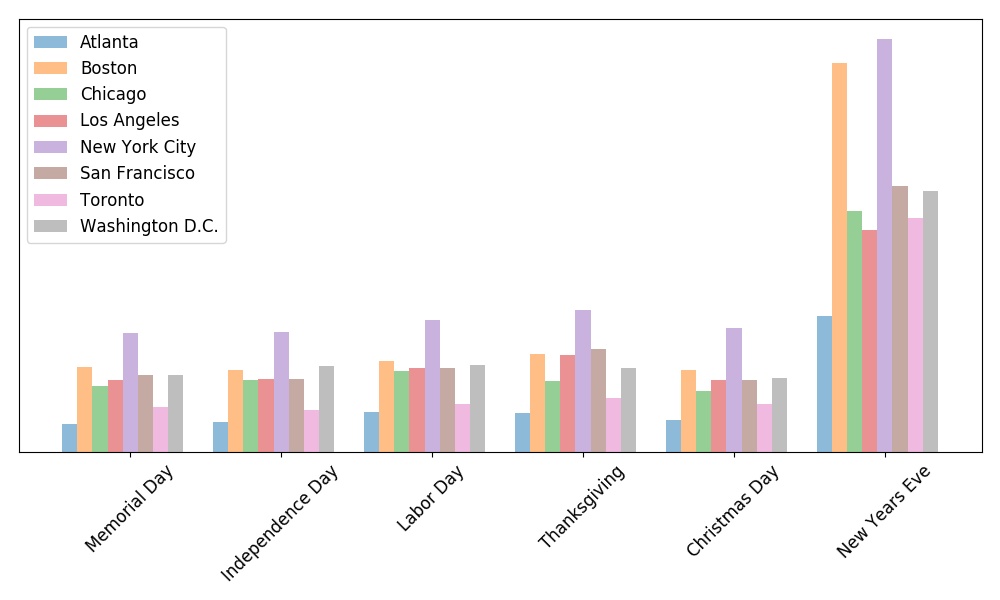}
\caption{Estimated prediction standard deviations on six U.S. holidays during testing period for eight cities. Exact values are anonymized. }
\label{fig:barplot-holiday}
\end{figure}

One important use-case of the uncertainty estimation is to provide insight for unusual patterns in the time series. \figurename~\ref{fig:barplot-holiday} shows the estimated predictive uncertainty on six U.S. holidays in the testing data. We see that New Year's Eve has significantly higher uncertainty than all other holidays. This pattern is consistent with our previous experience, where New Year's Eve is usually the most difficult day to predict. 

\subsubsection{Embedding features}
\begin{figure}[!t]
\centering
\includegraphics[width=0.5\textwidth]{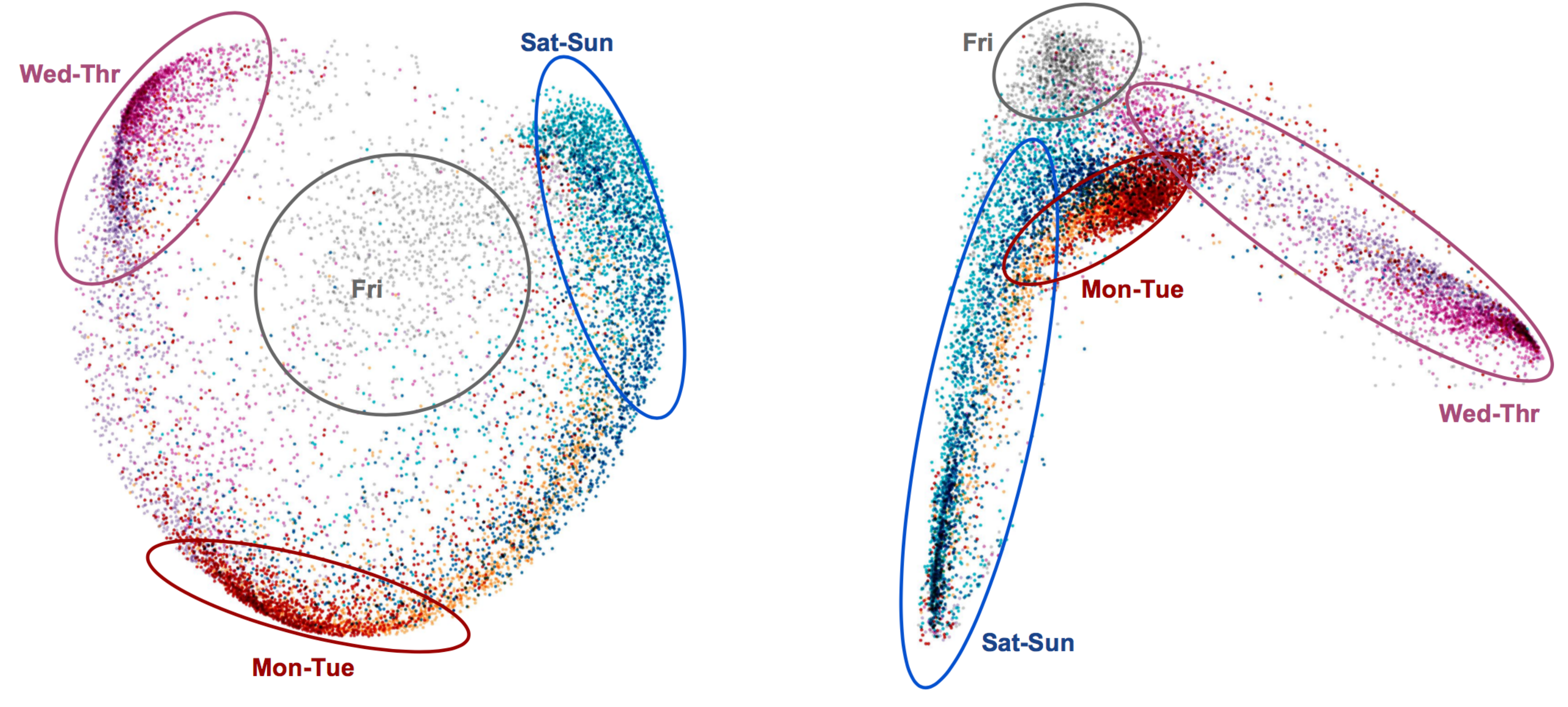}
\caption{Training set of time series, visualized in the embedding space. Each point represents a 28-day segment, colored by the day of the week of the last day. We evaluate the cell states of the two LSTM layers, where the first layer with dimension 128 is plotted on the left, and second layer with dimension 32 is plotted on the right. PCA is used to project into 2D space for visualization.}
\label{fig:embedding}
\end{figure}

As illustrated previously, the encoder is critical for both improving prediction accuracy, as well as for estimating prediction uncertainty. One natural follow-up question is whether we can interpret the embedding features extracted by the encoder. This can also provide valuable insights for model selection and anomaly detection. Here, we visualize our training data, each being a 28-day time series segment, in the embedding space. We use the last LSTM cell in the encoder, and project its cell states to 2D for visualization using PCA (\figurename~\ref{fig:embedding}). The strongest pattern we observe is day of the week, where weekdays and weekends form different clusters, with Fridays usually sitting in between. We do not observe city-level clusters, which is probably due to the fact all cities in this data set are large cities in North America, where riders and drivers tend to have similar behaviors.

\subsection{Application to Anomaly Detection 
at Uber
}

At Uber,
we track millions of metrics each day to monitor the status of various services across the company. One important application of uncertainty estimation is to provide real-time anomaly detection and deploy alerts for potential outages and unusual behaviors. A natural approach is to trigger an alarm when the observed value falls outside of the 95\% predictive interval. There are two main challenges we need to address in this application:
\begin{itemize}
\item Scalability: In order to provide real-time anomaly detection at the current scale, each predictive interval must be calculated within a few milliseconds during inference stage. 
\item Performance: With highly imbalanced data, we aim to reduce the false positive rate as much as possible to avoid unnecessary on-call duties, while making sure the false negative rate is properly controlled so that real outages will be captured.
\end{itemize}

\subsubsection{Scalability}
Our model inference is implemented in Go. Our implementation involves efficient matrix manipulation operations, as well as stochastic dropout by randomly setting hidden units to zero with pre-specified probability. A few hundred stochastic passes are executed to calculate the prediction uncertainty, which is updated every few minutes for each metric. We find that the uncertainty estimation step adds only a small amount of computation overhead and can be conducted within ten milliseconds per metric.

\subsubsection{Performance}
\begin{figure}[!t]
\centering
\subfloat[Normal I]{\includegraphics[width=0.25\textwidth]{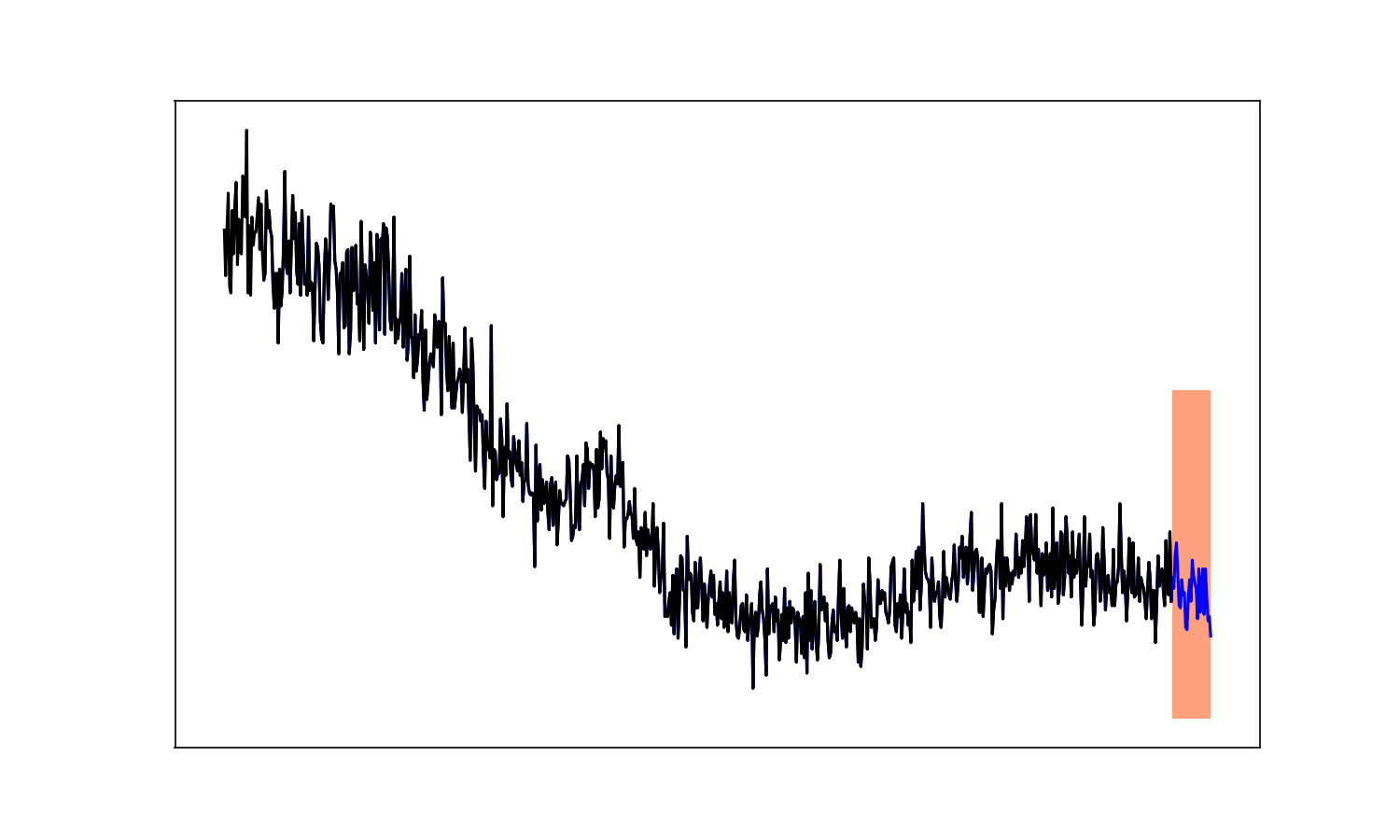}
\label{fig:normal-1}}
\subfloat[Normal II]{\includegraphics[width=0.25\textwidth]{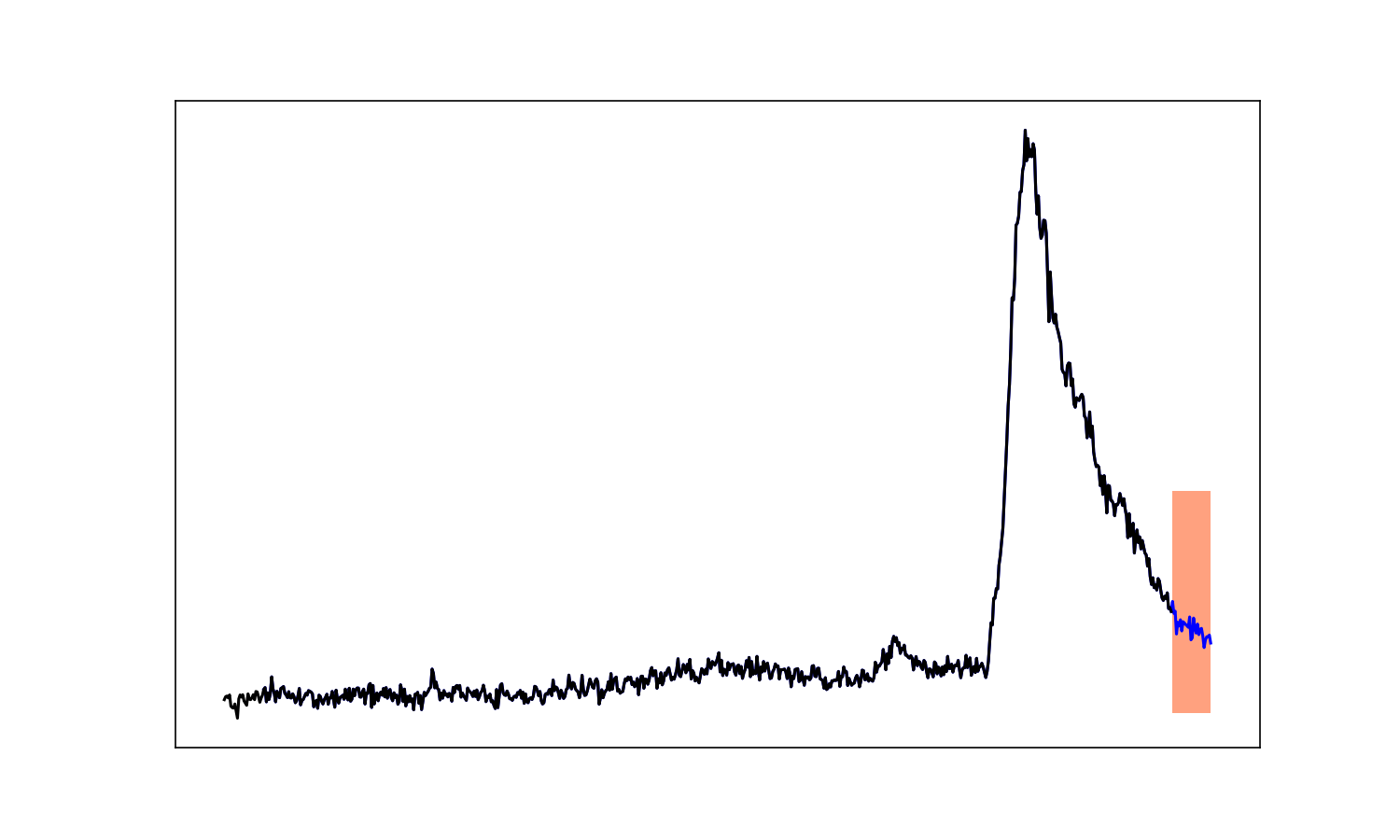}
\label{fig:normal-2}}

\subfloat[Anomaly I]{\includegraphics[width=0.25\textwidth]{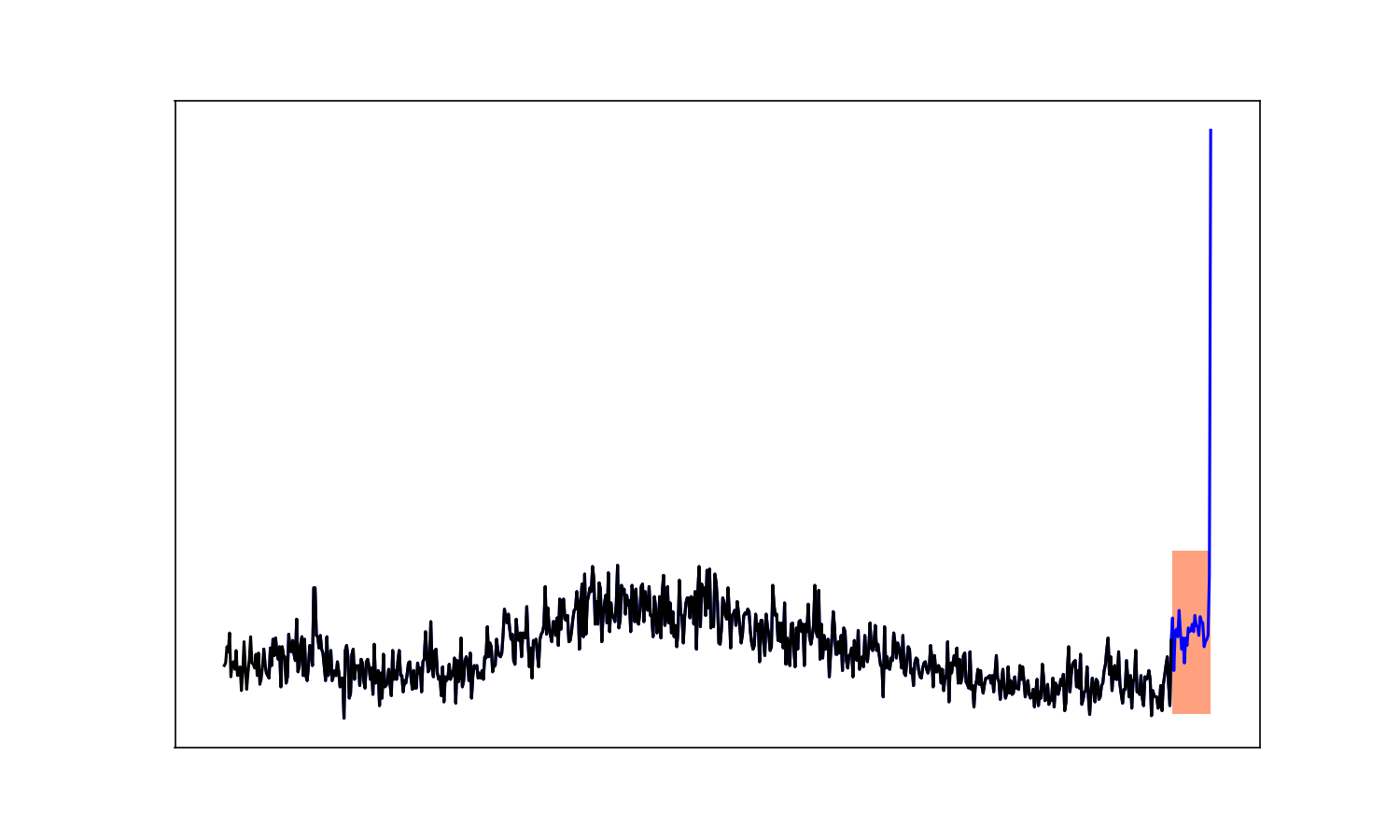}
\label{fig:anomaly-1}}
\subfloat[Anomaly II]{\includegraphics[width=0.25\textwidth]{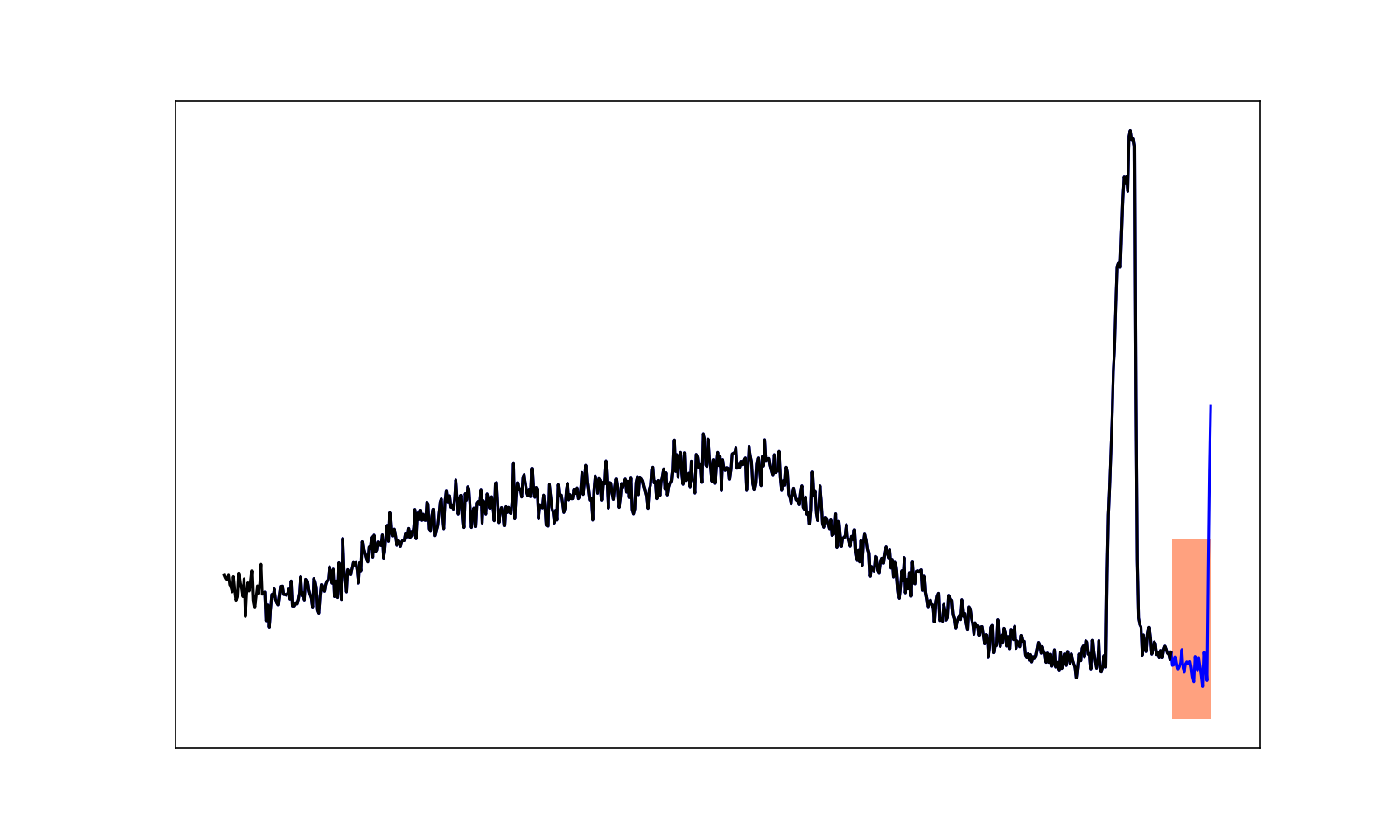}
\label{fig:anomaly-2}}

\caption{Four example metrics during a 12-hour span, and anomaly detection is performed for the following 30 minutes. All metrics are evaluated by minutes. The neural network constructs predictive intervals for the following 30 minutes, visualized by the shaded area in each plot. 
{\bf (a)} A normal metric with large fluctuation, where the observation falls within the predictive interval.
{\bf (b)} A normal metric with small fluctuation, and an unusual inflation has just ended. The predictive interval still captures the observation.
{\bf (c)} An anomalous metric with a single spike that falls outside the predictive interval.
{\bf (d)} An anomalous metric with two consecutive spikes, also captured by our model.
}
\label{fig:anomaly}
\end{figure}

Here, we illustrate the precision and recall of this framework on an example data set containing 100 metrics with manual annotation available, where 17 of them are true anomalies. Note that the neural network was previously trained on a separate and much larger data set. By adding MC dropout layers in the neural network, the estimated predictive intervals achieved 100\% recall rate and a 80.95\% precision rate. \figurename~\ref{fig:anomaly} visualizes the neural network predictive intervals on four representative metrics, where alerts are correctly fired for two of them. When applying this framework to all metrics, we observe a 4\% improvement in precision compared to the previous ad-hoc solution, which is substantial at Uber's scale.

\section{Conclusion}
\label{sec:conclusion}

We have presented an end-to-end neural network architecture for uncertainty estimation used 
at Uber. 
Using the MC dropout technique and model misspecification distribution, we showed a simple way to provide uncertainty estimation for a neural network forecast at scale while providing a 95\% uncertainty coverage. A critical feature about our framework is its applicability to any neural network without modifying the underlying architecture.

We have used the proposed uncertainty estimate to measure special event (e.g., holiday) uncertainty and to improve anomaly detection accuracy. For special event uncertainty estimation, we found New Year's Eve to be the most uncertain time. Using the uncertainty information, we adjusted the confidence bands of an internal anomaly detection model to improve precision during high uncertainty events, resulting in a 4\% accuracy improvement, which is large given the number of metrics we track 
at Uber. 

Our future work will be focused on utilizing the uncertainty information for neural network debugging during high error periods.

\bibliographystyle{IEEEtran}
\bibliography{BNN}

\end{document}